%% file: acl2023.tex
\pdfoutput=1
\documentclass[11pt]{article}

\usepackage[]{ACL2023}
\usepackage{times}
\usepackage{latexsym}
\usepackage[T1]{fontenc}
\usepackage[utf8]{inputenc}

\usepackage{microtype}

\usepackage{inconsolata}
\usepackage{xcolor}
\usepackage{multirow}
\usepackage{booktabs}
\usepackage{makecell}
\usepackage{amssymb}
\usepackage{colortbl}
\usepackage{graphicx}
\usepackage{bbding}
\usepackage{utfsym}
\usepackage{caption}
\usepackage{amsmath}
\usepackage{enumitem}

\usepackage[export]{adjustbox}

\definecolor{ao(english)}{rgb}{0.0, 0.5, 0.0}
\definecolor{darkgreen}{rgb}{0.0, 0.5, 0.0}
\usepackage{pifont}
\usepackage{xcolor}
\usepackage{xspace}
\definecolor{ForestGreen}{RGB}{34,139,34}
\definecolor{BrickRed}{rgb}{.72,0,0}
\definecolor{LakeBlue}{RGB}{0,61,153}
\usepackage{placeins}

\usepackage{tcolorbox}
\newtcolorbox{finding}{
colframe=black!80,
colback=icyrockblue,
fonttitle=\bfseries,
coltitle=black,
left=3pt,
right=3pt,
top=3pt,
bottom=3pt,
boxrule=0.4mm,
arc=3mm
}

\usepackage[utf8]{inputenc} % allow utf-8 input
\usepackage[T1]{fontenc}    % use 8-bit T1 fonts
\usepackage{hyperref}       % hyperlinks
\usepackage{url}            % simple URL typesetting
\usepackage{booktabs}       % professional-quality tables
\usepackage{amsfonts}       % blackboard math symbols
\usepackage{nicefrac}       % compact symbols for 1/2, etc.
\usepackage{microtype}      % microtypography
\usepackage{xcolor}         % colors
\usepackage{wrapfig}

\usepackage{algorithm}
\usepackage{algorithmicx}
\usepackage{algpseudocode}

\usepackage{inconsolata}
\usepackage{xcolor}
\usepackage{multirow}
\usepackage{booktabs}
\usepackage{makecell}
\usepackage{amssymb}
\usepackage{colortbl}
\usepackage{graphicx}
\usepackage{bbding}
\usepackage{utfsym}
\usepackage{caption}
\usepackage{pifont}
\usepackage{amsmath}
\usepackage{enumitem}
\usepackage[export]{adjustbox}% http://ctan.org/pkg/adjustbox
\usepackage{xspace}
\usepackage{mathrsfs}

\definecolor{ForestGreen}{RGB}{34,139,34}
\definecolor{BrickRed}{rgb}{.72,0,0}
\definecolor{LakeBlue}{RGB}{0,61,153}

\definecolor{self1}{RGB}{192,0,0}
\definecolor{self2}{RGB}{46,117,182}
\definecolor{self3}{RGB}{118,113,113}

\definecolor{ao(english)}{rgb}{0.0, 0.5, 0.0}
\definecolor{veronica-red}{RGB}{196,30,58}

\newcommand{\ours}{\textsl{\(\phi\)-Decoding}\xspace}
\usepackage{fontawesome}
\definecolor{ForestGreen}{RGB}{34,139,34}
\definecolor{BrickRed}{rgb}{.72,0,0}
\definecolor{LakeBlue}{RGB}{0,61,153}

\definecolor{strings}{RGB}{220, 20, 60}

\author{
Fangzhi Xu\textsuperscript{2,1}\thanks{\, means equal contribution. Work done during Fangzhi's internship at Shanghai AI Lab.} \quad
Hang Yan\textsuperscript{2*} \quad
Chang Ma\textsuperscript{3} \quad
Haiteng Zhao\textsuperscript{4} \quad \\
\bf{
Jun Liu\textsuperscript{2}\footnotemark[2] \quad
Qika Lin\textsuperscript{5}\footnotemark[2] \quad
Zhiyong Wu\textsuperscript{1}\thanks{\, denotes corresponding author.} \quad
}\\
\textsuperscript{1}Shanghai AI Lab \quad
\textsuperscript{2}Xi'an Jiaotong University \quad
\textsuperscript{3}The University of Hong Kong \quad \\
\textsuperscript{4}Peking University \quad
\textsuperscript{5}National University of Singapore \quad \\
\texttt{\{fangzhixu98,whucs2013wzy\}@gmail.com} \:
\texttt{hangyan666@outlook.com} \:
\texttt{cma@cs.hku.hk} \quad \\
\texttt{zhaohaiteng@pku.edu.cn} \quad
\texttt{liukeen@xjtu.edu.cn} \quad
\texttt{linqika@nus.edu.sg}
\\
}

\title{$\phi$-Decoding: Ada\underline{p}tive Foresig\underline{h}t Sampl\underline{i}ng for Balanced Inference-Time Exploration and Exploitation}

\begin{document}
\maketitle

\input{sections/0.abstract}
\input{sections/1.Introduction_modified}
\input{sections/3.approach}
\input{sections/4.experiments}

\input{sections/5.analysis}
\input{sections/2.relatedwork}

\input{sections/6.conclusion}

% Entries for the entire Anthology, followed by custom entries
% \bibliography{anthology,custom}
\bibliography{custom}
\bibliographystyle{acl_natbib}

\clearpage
\newpage
\appendix

\input{sections/appendix}

\end{document}

%% file: sections/0.abstract.tex
\begin{abstract}
Inference-time optimization scales computation to derive deliberate reasoning steps for effective performance. 
While previous search-based strategies address the short-sightedness of auto-regressive generation,
% they either face \xfz{xxx} large search spaces or inadequate step value assessments.
the vast search space leads to excessive \emph{exploration} and insufficient \emph{exploitation}.
% \xfz{xxx} To balance exploration and exploitation during inference,
To strike an efficient balance to derive the optimal step,
we frame the decoding strategy as \emph{foresight sampling}, 
% \xfz{explain the function of step value}
leveraging simulated future steps to obtain globally optimal step estimation.
Built on it, we propose a novel decoding strategy, named \ours.
% \xfz{propose a novel approach to estimate step value}
% (1) \emph{how to estimate step value with the foresight steps?} and (2) \emph{whether every step requires deliberation for decision-making?}
To provide a precise and expressive estimation of step value,
\ours approximates two distributions via foresight and clustering.
Sampling from the joint distribution, the optimal steps can be selected for exploitation.
To support adaptive computation allocation, we propose in-width and in-depth pruning strategies,
featuring a light-weight solution to achieve inference efficiency.
Extensive experiments across seven benchmarks show \ours outperforms strong baselines in both performance and efficiency. Additional analysis demonstrates its generalization across various LLMs and scalability across a wide range of computing budgets.\footnote{The code will be released at \url{https://github.com/xufangzhi/phi-Decoding}, and the open-source PyPI package is coming soon.}

\end{abstract}

%% file: sections/1.Introduction_modified.tex
\section{Introduction}

Large language models (LLMs)~\cite{achiam2023gpt, team2023gemini} present remarkable performances in solving reasoning-intensive tasks through step-by-step thoughts~\cite{wei2022chain}.
Recent advancements~\cite{qwq-32b-preview, guo2025deepseek} have significantly boosted LLM reasoning by large-scale post-training on well-curated datasets.
% Nevertheless, the huge cost associated with the post-training procedure hinders its reproducibility. \changh{Not sure post-training has huge cost, maybe change to:}\chang{Recent result show that inference-time strategy could be superior for optimizing LLM reasoning at a lower cost \citep{}}
Nevertheless, the cost associated with the post-training procedure hinders its reproducibility.
This naturally motivates us to explore the inference-time strategy for optimizing the LLM reasoning chains.

\begin{figure}[t]
    \centering
    \includegraphics[scale=0.42]{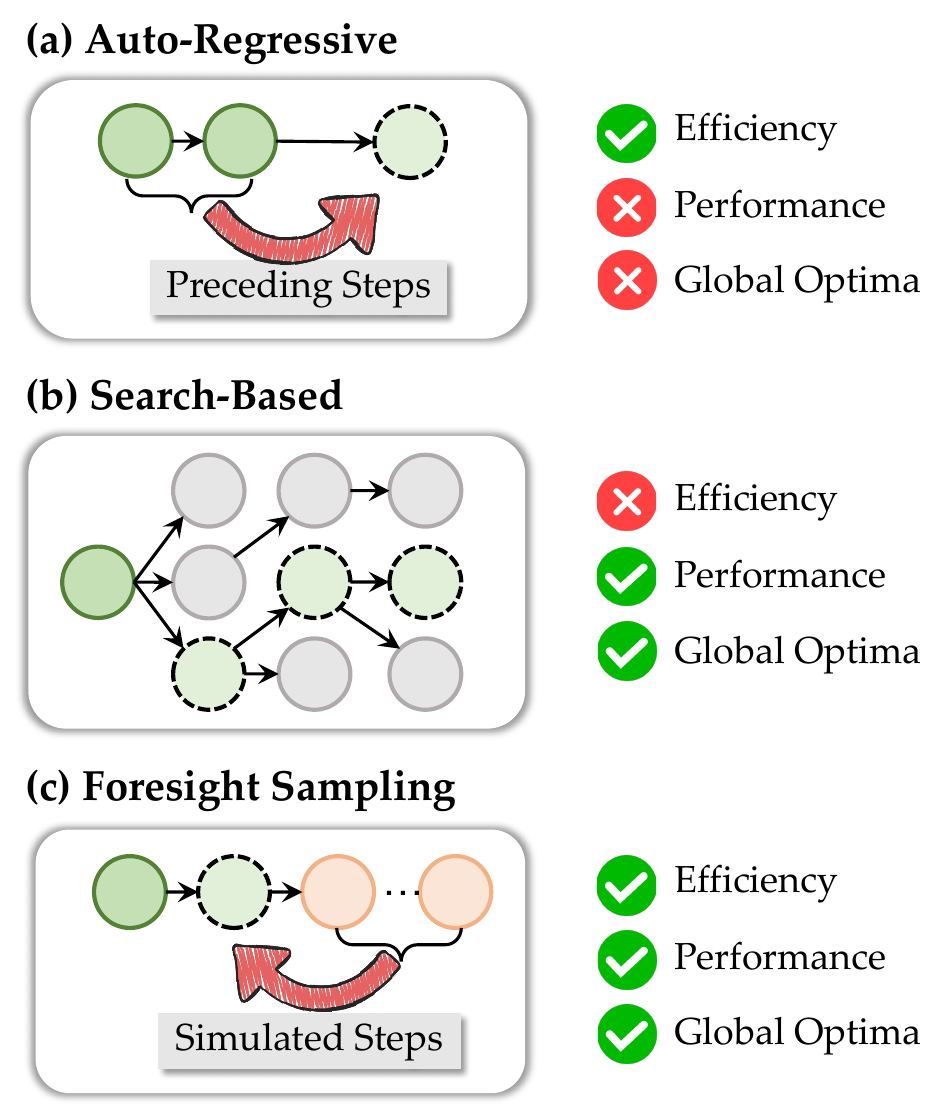}
    \vspace{-0.1cm}
    \caption{Comparisons of different decoding paradigms. (a) is auto-regressive decoding, which has high efficiency but lacks global awareness. (b) represents search-based methods, which requires huge search space with extensive time cost. (c) is the foresight sampling strategy. It leverages the simulated future steps to estimate the step value, which can strike a balanced inference-time exploration and exploitation.}
    \label{fig:foresight_sampling}
\end{figure}

Inference-time optimization involves employing more reasoning tokens that encode thinking steps to perform effective reasoning. 
However, the natural shortsightedness of auto-regressive generation, which predicts the next step only with preceding steps, makes most inference algorithms unable to achieve global optima~\cite{ma2024non} (Fig.~\ref{fig:foresight_sampling}(a)).
Most previous work solves this by deliberately optimizing each step using search-based methods~\citep{yao2024tree, hao2023reasoning, xie2024self,wu2024inference}, the expanding and backtracking of tree search algorithms enable LLMs to find global-optimal reasoning paths (Fig.~\ref{fig:foresight_sampling}(b)). 
However, the vast search space results in excessive \emph{exploration} and insufficient \emph{exploitation}.
% However, these approaches often entail extensive time costs and huge search space. 
% In contrast, if we could have a precise estimation of global-aware step values, less exploration is necessary and could improve the efficiency of reasoning~\citep{wang2024math, ma2024non}. Therefore, to strike an efficient balance between inference-time exploration and exploitation, it is inevitable to obtain reliable and expressive step value estimations.
Conversely, if we could derive a precise estimation of globally-aware step values,
an efficient balance between inference-time exploration and exploitation could be achieved.
% the exploration cost could be reduced while enhancing the reasoning efficiency.
% Therefore, the objective of our effort is to \emph{strike an efficient balance between inference-time exploration and exploitation}.

% Based on this objective of \emph{striking an efficient balance between inference-time exploration and exploitation},
Based on this,
we frame the decoding strategy as \textbf{\emph{foresight sampling}}, as depicted in Fig.~\ref{fig:foresight_sampling}(c).
It relies on the future simulation to obtain the globally optimal estimation of the current step.
% In Figure~\ref{fig:foresight_sampling}, we illustrate the comparison between decoding paradigms.
Central to the foresight sampling is the critical task of \emph{how to estimate step value with the foresight steps}.
% Without external reward models, the step values can be estimated either through self-prompting~\cite{yao2024tree, hao2023reasoning, xie2024self}, or model uncertainty~\cite{ma2024non}.
Intuitively, the step estimation with foresight can be derived either by incorporating the process reward model (PRM)~\cite{snell2024scaling} or through model uncertainty~\cite{ma2024non}.
However, PRMs are not widely available for all reasoning scenarios, which poses challenges for scalability.
Delegating the step assessment to model uncertainty risks the issue of local optima, potentially resulting in suboptimal performance.

% Fig.~\ref{fig:step_value} shows the performances on the step value estimation accuracy as well as the task performance.
% It reveals that there is still significant room for improvement in previous strategies,
% underscoring our method as the optimal selection.

Another issue in stepwise exploration and exploitation is \emph{whether every step requires deliberation for decision-making}.
Naturally, more computational resources should be allocated to challenging steps, while conserving compute for simpler steps.
% Fig.X(c) gives a pilot observation that cutting xxx \% of thinking computation would only lead to xxx\% performance drops.\xfz{pilot exp}
Previous inference-time optimization methods widely overlook this principle.
In addition, such concept of \emph{over-thinking} has been widely observed in the o1-like attempts~\cite{chen2024not,manvi2024adaptive}.
% However, few efforts have been made to alleviate the unnecessary compute.
Therefore, it is both intriguing and promising to develop a light-weight solution that can adaptively balances computational workload without extra training.

In this paper, we propose a novel inference-time optimization algorithm named \ours, which introduces an adaptive foresight sampling strategy to achieve efficient exploration and exploitation during inference.
To give the reliable and expressive step value estimation, 
\ours capitalizes on foresight paths to derive two distributions: one from the derived step \emph{advantage} values, capturing uncertainty discrepancies between successive steps, and another from \emph{alignment} of these foresight paths via clustering. 
Sampling from the joint distribution, \ours selects optimal steps for exploitation.
To efficiently allocate the computations,
\ours introduces both the in-width and in-depth pruning strategies,
which provides adaptive inference-time scaling.

On diverse reasoning benchmarks,
\ours improves the average performance of LLaMA3.1-Instruct-8B by >14\% over auto-regressive CoT.
Inference-time scaling across diverse computing budgets shows the consistent superiority of \ours over other baselines,
offering a balance between performance (Accuracy) and computational efficiency (\#FLOPS). 
% We also provide insightful \changh{be more specific, and mention conclusions here} analyses of \ours's generalizability to other backbone LLMs and the underlying reasons for its superiority.
Further analysis of the generalization across various backbone LLMs and scalability to the competition-level task highlights the superiority of \ours.

The major contributions of our work are:

\noindent (1) \textbf{An adaptive inference-time optimization algorithm \ours without external auxiliary}. \ours estimates the step value based on the joint distribution derived from foresight paths. In-width and in-depth pruning strategies are introduced to alleviate the overthinking issue.

\noindent (2) \textbf{Extensive experiments with state-of-the-art performances.}
\ours improves the average reasoning of LLaMA3.1-8B-Instruct by over 14\% across various reasoning benchmarks, presenting a great trade-off between effectiveness and efficiency compared with baselines.

% Its generalizability to other backbone LLMs is well verified.

% \noindent (3) \textbf{Insightful findings and discussions.}\changh{This title of this point should be more specific, like mentioning the most key insightful finding}
% Our research also provides valuable insights into the underlying reasons for superiority as well as the scaling of inference-time computing,
% paving the way for future research.

\noindent (3) \textbf{Comprehensive analysis and insightful findings.}
\ours proves its generalization ability to various LLMs, ranging from the 70B-sized model to R1-distilled LLM.
Additionally, the inference-time scaling across a wide range of computing budgets reveals the consistent advantages, where \ours matches the performance of the suboptimal baseline with 6$\times$ efficiency.

%% file: sections/3.approach.tex
 \section{Methodology}

\begin{figure*}[t]
\large
\centering
\includegraphics[scale=0.45]{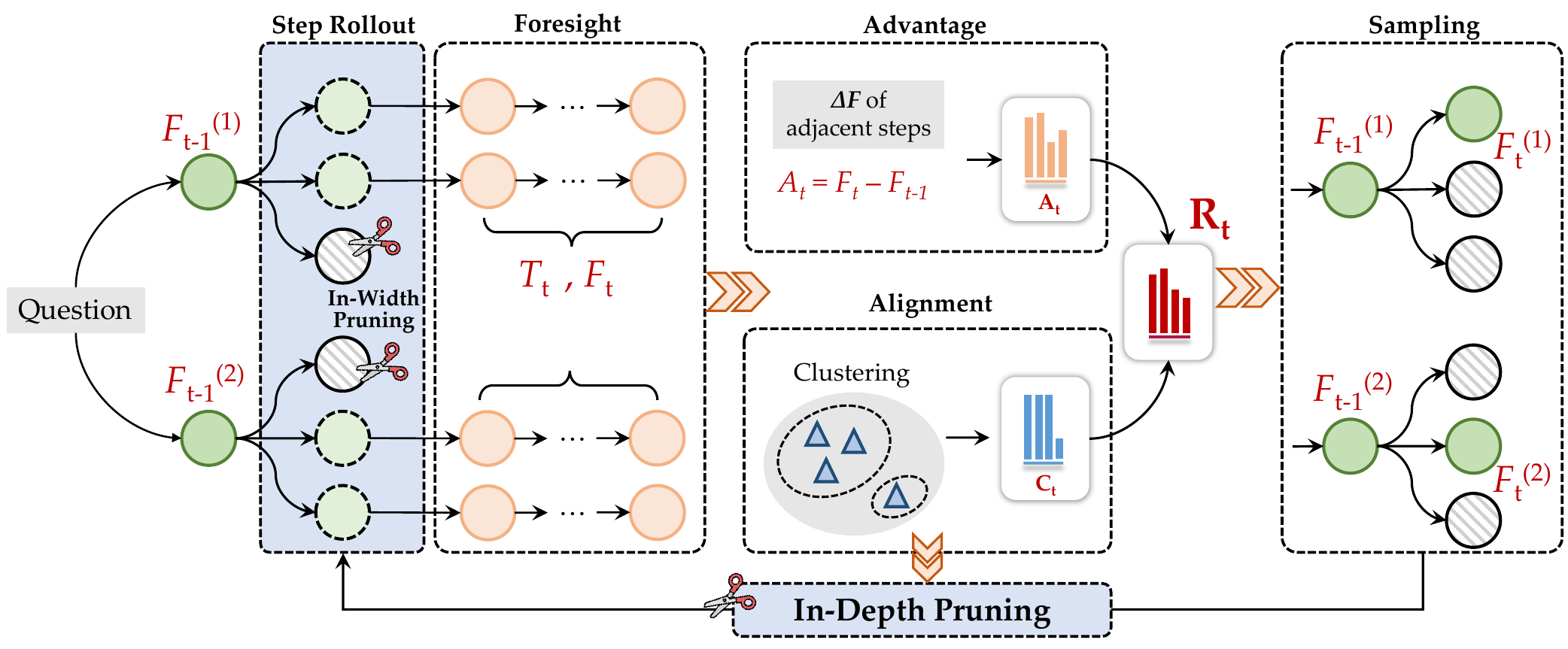}
\caption{The overall framework of \ours. We visualize the decoding process at the timestamp $t$. For simplicity, we set step beam size $M$ as 2, the number of rollouts $N$ as 3, and the number of clusters $K$ as 2.}
\label{model}
% \vspace{-1em}
\end{figure*}

\subsection{Preliminary}

In the context of auto-regressive language generation,
the selection of the current step $\hat{a}_t$ is based on the following probability distribution:

\begin{equation}
    \hat{a}_t \sim p_\theta(a_t | x,\mathbf{a}_{<t})
\end{equation}
% \changh{add $a_t$ to the bracket}
where $x$ is the instruction or the input query, and $\textbf{a}_{<t}$ represents the preceding steps.
$\theta$ denotes the LLM parameters, where $p_\theta$ is derived from the distribution of language modeling.

To overcome the short-sighted limitation of auto-regressive generation and achieve efficient exploration,
\emph{foresight sampling} conditions the generation process not only on the preceding steps $\textbf{a}_{<t}$ but also on an estimation of future outcomes $\textbf{a}_{>t}$. 
We use the Boltzmann distribution to model the probabilities of different outcomes during the decoding process,
incorporating both the influence of preceding steps and an estimation of future states, such as: 
% the \emph{foresight sampling} is formulated as\changh{add more description here on how to overcome short-sightedness: Foresight sampling conditions the generation process not only on the preceding steps but also on an estimation of future outcomes. We use the Boltzmann distribution to model the probabilities of different outcomes during the decoding process, incorporating both the influence of preceding steps and an estimation of future states, such as: }
\begin{equation}
\label{eq_foresight}
    \hat{a}_t \sim p_\theta(a_t |x,\mathbf{a}_{<t}) \mathbb{E}_{\mathbf{a}_{>t}} p_\theta(\mathbf{a}_{>t} |x, a_t, \mathbf{a}_{<t}) 
\end{equation}

% \changh{It is non-trivial to precisely calculate $\mathbb{E}_{\mathbf{a}_{>t}} p_\theta(\mathbf{a}_{>t} |x, a_t, \mathbf{a}_{<t})$. Therefore, we try to obtain an estimation of this future simulation quality.}
It is non-trivial to have a precise calculation of $\mathbb{E}_{\mathbf{a}_{>t}} p_\theta(\mathbf{a}_{>t} |x, a_t, \mathbf{a}_{<t})$. 
Therefore, we try to derive an estimation of this future simulation quality.
% In this standard formulation, the current step is determined by both previous steps ($\mathbf{a}_{<t}$) and future simulation ($\mathbf{a}_{>t}$).
% For better estimation of the step value, it is required to obtain the optimal distribution of the step, which is formed as:
\begin{equation}
\label{eq_objective}
    \hat{a}_t \sim p_\theta(a_t|x,\mathbf{a}_{<t}) \mathrm{exp} \left[ R(x, \mathbf{a}_{\leq t}, \mathbf{a}_{>t}) / \tau \right]
\end{equation}
where $R$ denotes the optimized function for step value estimation based on the foresight steps.
$\tau$ represents the temperature hyper-parameter, which controls the diversity of generation.

Therefore, the ultimate objective of \ours is to design the step value estimation function $R(x, \mathbf{a}_{\leq t}, \mathbf{a}_{>t})$.
We include the key techniques of \ours in Fig.~\ref{model}, which depicts the decoding process at the timestamp $t$. 
The complete algorithm as well as the overall decoding pipeline are presented in Appendix~\ref{ref:algo}.

\subsection{Step Value Estimation}

% From the standard formulation of foresight sampling (Eq.~\ref{eq_foresight}),
% the $R$ function is simplified by the model uncertainty of the foresight steps.
To thoroughly optimize the formulation,
we propose to evaluate the foresight paths from \emph{advantage} (absolute value) and \emph{alignment} (relative value).
% Figure~\ref{model}(a) shows the core idea of the step value estimation method.

\paragraph{Dynamic Advantage Estimation.}
\label{sec_adv}
We follow the beam search strategy.
% At Step $t-1$, we keep $M$ beams, denoted as $\{a_{t-1}^{(m)} | m \in [0, M-1]\}$. 
% The corresponding step values are represented as $Q_{t-1}^{(m)}$.
% At the timestamp $t-1$, the updated value of the last step $a_{t-1}$ in each beam can be written as $Q_{t-1}$.
% For simplicity, we omit the beam index in the symbol.
% In Fig.~\ref{model}(a), the step beam size is set as 2 for illustration.
At the timestamp $t$, we rollout $N$ candidate steps from each beam.
Based on the idea of \emph{foresight sampling}, the probability $F_t$ of the foresight path can be derived:
\begin{equation}
    F_t = p_\theta(\mathbf{a}_{>t} |x, a_t, \mathbf{a}_{<t}),
\end{equation}
where the index of the candidate step is omitted for simplicity.
% Then, the value of $a_t$ is updated with $F_t$, formulated as $Q_t := F_t$.
% That is, we leverage the uncertainty of the foresight path to measure the 

To measure the advantage brought by the candidate step $a_t$, we define the calculation of \emph{Advantage} $A_t$ as:
\begin{equation}
    \begin{aligned}
        A_t &= p_\theta(\mathbf{a}_{>t} |x, a_t, \mathbf{a}_{\le t}) - p_\theta(\mathbf{a}_{>t-1} |x, a_{t-1}, \mathbf{a}_{< t-1}) \\
        &= F_t - F_{t-1}
    \end{aligned}
\end{equation}

% \hangyan{the equation above makes me feel like $\mathbf{a}_{>t}$ is part of $\mathbf{a}_{>t-1}$. eg: $\mathbf{a}_{\leq t}$ is 1234, $\mathbf{a}_{>t}$ is 567, $\mathbf{a}_{\leq t-1})$ is 123,$\mathbf{a}_{>t-1}$ is 4567. In fact, if $\mathbf{a}_{>t}$ is 567, $\mathbf{a}_{>t-1}$ can be 6666}

It is represented as the $\Delta$ of the foresight probability $F$ between the adjacent steps.
Notably, we implement the calculation of $p_\theta$ with the averaged log probability of the sequence, 
which alleviates the influence from the foresight length.

Since the calculation of \emph{Advantage} for each candidate step is independent,
it estimates the absolute value of the step.
For better illustration, we define $R_1(x, \mathbf{a}_{\leq t}, \mathbf{a}_{>t}) = \mathrm{exp}(A_t / \tau_1)$.

\paragraph{Alignment Assessment by Clustering.}
\label{sec_align}
One potential risk of the uncertainty-based estimation is the issue of local optima.
That is, LLMs may be trapped in the incorrect step with exceptionally high confidence.

To address this limitation, we introduce the definition of \emph{alignment} to provide the relative preference among the foresight paths.
This is achieved by employing a clustering strategy following the foresight sampling process.
Specifically, the foresight paths at each timestamp are grouped into clusters.
The number of clusters is defined as $K$.
The \emph{alignment} value of $a_t$ is determined based on the size of the cluster to which it belongs:

\begin{equation}
    C_t = \frac{|\textrm{Cluster}(a_t)|}{\# \textrm{Foresight Paths}}
\end{equation}
where $|\textrm{Cluster}(a_t)|$ denotes the size of the cluster $a_t$ belongs to.

\emph{Alignment} actually provides the relative estimation of the step value, which reflects the consistency among the foresight paths.
The more closely the expected outcome aligns with those of other candidates, the greater the step value would be.
Similarly, we define $R_2(x, \mathbf{a}_{\leq t}, \mathbf{a}_{>t}) = \mathrm{exp}(C_t / \tau_2)$.

\paragraph{Sampling From Joint Distribution}

Combining the rewarding from $R_1$ and $R_2$,
we can derive the definition of $R$ function, which is in the form of:
\begin{equation}
    \begin{aligned}
    R(x, \mathbf{a}_{\leq t}, \mathbf{a}_{>t}) = \mathrm{Norm}(A_t) + \mathrm{Norm}(C_t) \\
    = \frac{\mathrm{exp}(A_t / \tau_1)}{\sum_{a_t} \mathrm{exp}(A_t / \tau_1)} + \frac{\mathrm{exp}(C_t / \tau_2)}{\sum_{a_t} \mathrm{exp}(C_t / \tau_2)}
    \end{aligned}
\end{equation}

Replacing this formulation of $R$ into Eq.~\ref{eq_objective},
the objective becomes the sampling on the joint distribution of \emph{Advantage} and \emph{Alignment}.

In the implementation, we set $\tau_1=\tau_2=0.6$ and combine $R_1$ and $R_2$ with equal weighting for simplicity.
We leave the discussion of the weighted version in future work.
% \hangyan{$\tau_1, \tau_2$ are the same, you seem not to mention it. (Mentioning it may reduce the complexity of our hyperparameters)}

\subsection{Dynamic Pruning Strategy}

To optimize the computation allocation and alleviate the over-thinking issue,
we introduce an efficient and effective pruning strategy.
It is designed from two dimensions: in-width and in-depth.
Figure~\ref{model} visualizes the function of the pruning stratgies.

\paragraph{In-Width Pruning.}
Although foresight sampling addresses the short-sightedness of language models, it inevitably introduces additional computational cost. 
Intuitively, some steps with obvious errors can be filtered out directly, without needing to simulate future steps. 
To achieve this, we assess the generation confidence of each step $a_t$ based on its probability:

\begin{equation}
    s_t = p_\theta(a_t |x,\mathbf{a}_{<t}).
\end{equation}
% For each candidate step $i$ 
% \hangyan{may be easier to understand with: there are $M*N$ candidate steps in total}
There are in total $M*N$ candidate steps at timestamp $t$.
We then calculate the mean and variance of these step confidence:
\begin{equation}
    \mu_t = \frac{1}{M*N} \sum_{i} s_t^{(i)}, \, \, \sigma_t^2 = \frac{1}{M*N} \sum_{i} (s_t^{(i)} - \mu)^2
\end{equation}
% \hangyan{In fact, we use $\sigma_t=\sqrt{  \frac{1}{M*N} \sum_{i} (s_t^{(i)} - \mu)^2}$ to do in-width pruning, consider to arrange $\mu_t$ and $\sigma_t$ in two rows if it's too long to write them in one row}

% \changh{think you need to add $t$ to $\mu$ and $\sigma$, i.e. $\mu_t$, the same goes for this paragraph}
where $\mu$ and $\sigma^2$ denote the mean and variance values respectively.
$M*N$ is the number of candidates under the setting of step beam search as defined in Sec.~\ref{sec_adv}.

Based on this calculation, we exclude any steps with generation confidence that is exceptionally low, i.e., those with $s_t^{(i)} < \mu - \sigma$.
The remaining steps are kept for foresight:
\begin{equation}
    \mathscr{S}_t = \{a_t^{(i)} | \mu - \sigma \leq s_t^{(i)} \}
\end{equation}

Adhering to this principle enables the attainment of in-width pruning. 
Notably, the extent of pruning can be regulated by adjusting the threshold using $\mu - k\sigma$, where $k \in Z^{+}$. 

% \hangyan{For simplicity, we adopt $k=1$ here.} 

% \hangyan{in fact, $k$ can be any digit here, (not only $Z^{+}$ }
% \changh{range/ scale of $k$, or just a real number, or the normal number you set for experiments.}

\paragraph{In-Depth Pruning.}
Foresight sampling enables the deliberate thinking of each step.
Previous work~\cite{wang2024chain} uncovers that the early steps are much more important, necessitating increased computational resources for optimization.
As the final answer approaches, LLMs exhibit greater determination in their reasoning paths.
Motivated by it, we can save some computational costs with the strategy of early stopping.

To avoid extra computing and make the solution as simple as possible,
we employ the clustering result introduced in Sec.~\ref{sec_align}.
In detail, we derive the size of the largest cluster, written as $|\textrm{Cluster}_{max}|$.
The condition of early-stopping is controlled by the threshold:
\begin{equation}
    \frac{|\textrm{Cluster}_{max}|}{\# \textrm{Foresight Paths}} \geq \delta
\end{equation}
Then, the LLM completes the remaining reasoning steps under the auto-regressive setting.
For convenience, we set $\delta=0.7$ for all experiments.

%% file: sections/4.experiments.tex
\section{Experiments}

\subsection{Evaluation Benchmarks and Metrics} 

\paragraph{Benchmarks}
To comprehensively evaluate the LLM performances on downstream tasks, we mainly include the following 6 representative reasoning benchmarks:
GSM8K~\cite{cobbe2021training}, MATH-500~\cite{hendrycks2021measuring}, GPQA~\cite{rein2023gpqa}, ReClor~\cite{yureclor}, LogiQA~\cite{liu2021logiqa}, and ARC-Challenge~\cite{clark2018think}. 
Furthermore, we incorporate the competition-level benchmark AIME~\cite{aime_2024} to highlight the scalability of \ours to address more challenging scenarios.

\paragraph{Metrics}
We report the Pass@1 accuracy (Acc.) for each benchmark.
To better illustrate the trade-off between efficiency and performance, the FLOPS metric is also computed, following the definition of ~\cite{kaplan2020scaling}.
Please refer to Appendix~\ref{ref:flops} for more evaluation details.

\subsection{Baselines and Backbone LLMs}

In the experiments, we compare \ours with the following 5 baseline methods.

\paragraph{Auto-Regressive (CoT).} It produces the chain-of-thought reasoning through the auto-regressive language generation.

\paragraph{Tree-of-Thought (ToT)~\cite{yao2024tree}.}
It builds a tree structure for a given problem, where each node represents a reasoning step. 
We use the BFS version as the implementation.

\paragraph{Monte Carlo Tree Search (MCTS).} It constructs a search tree and dynamically updates the step value via expanding and backtracking.
We follow Reasoning as Planning (RaP)~\cite{hao2023reasoning} for implementation.

\paragraph{Guided Decoding~\cite{xie2024self}.}
It utilizes self-evaluation at each step to perform a stochastic beam search.

\paragraph{Predictive Decoding~\cite{ma2024non}.}
It proposes the look-ahead strategy and leverages Model Predictive Control to reweigh LLM distributions,
producing non-myopic language modeling.

For the 6 reasoning benchmarks in the main experiments, all the baseline methods are evaluated on two backbone LLMs: LLaMA3.1-8B-Instruct~\cite{dubey2024llama} and Mistral-v0.3-7B-Instruct~\cite{jiang2023mistral}.
To assess generalization and scalability, we further evaluate the Qwen2.5-3B~\cite{yang2024qwen2} and LLaMA3.1-70B~\cite{dubey2024llama} LLMs, while also boosting Deepseek R1-series LLM (i.e., R1-Distill-LLaMA-8B)~\cite{guo2025deepseek} for competition-level tasks.

All the experiments are implemented on A100 of 80GB VRAM GPUs.
The inference process is accelerated by the vLLM engine~\cite{kwon2023efficient}.
The generation temperature is set to 0.6.
% The temperature for each generation step is set to 1 as default.
Please refer to Appendix~\ref{ref:setup} for more implementation details.

\subsection{Main Results}

Table~\ref{exp_main} presents the results on 6 reasoning benchmarks across 2 representative open-source LLMs. 

\begin{table*}[t]
\centering
\footnotesize
\resizebox{\linewidth}{!}{
\begin{tabular}{l|cccccc|cr}
    \toprule
    \multirow{1}{*}{\textbf{Models}}  &\textbf{GSM8K} &\textbf{Math-500} &\textbf{GPQA} &\textbf{ReClor} &\textbf{LogiQA} &\textbf{ARC-c} 
 &\multirow{1}{*}{\textbf{Avg.}} &\multirow{1}{*}{\textbf{FLOPS}} \\
    \midrule
    \multicolumn{9}{c}{\cellcolor{gray!25} LLaMA3.1-8B-Instruct}  \\
    \midrule
    Auto-Regressive (CoT) &70.28 &31.00 &26.56 &49.40 &33.33 &58.91 &44.91 &$1.34\times10^{16}$\\
    % Best-of-N & \\
    % Beam Search & \\
    Tree-of-Thoughts &75.74 &31.60 &\underline{31.25} &59.00 &45.93 &80.72 &54.04 &$7.03\times10^{17}$ \\
    MCTS &80.44 &\underline{34.40} &24.11 &61.40 &42.70 &79.95 &53.83 & $17.90\times10^{17}$ \\
    Guided Decoding &75.51 &31.20 &30.58 &60.20 &43.47 &81.74 &53.78 &$6.54\times10^{17}$ \\
    Predictive Decoding &\underline{81.43} &34.00 &31.03 &\textbf{64.00} &\underline{46.70} &\underline{84.56} &\underline{56.95} &$6.89\times10^{17}$ \\
    \midrule
    \ours &\textbf{86.58} &\textbf{38.20} &\textbf{34.60} &\textbf{64.00} &\textbf{48.39} &\textbf{85.41}	 &\textbf{59.53} &$6.43\times10^{17}$ \\
    \midrule
    \multicolumn{9}{c}{\cellcolor{gray!25} Mistral-v0.3-7B-Instruct}  \\
    \midrule
    Auto-Regressive (CoT) &49.05 &\underline{12.20} &23.88 &52.20 &37.02 &69.54 &40.65 &$0.81\times10^{16}$ \\
    % Best-of-N & \\
    % Beam Search & \\
    Tree-of-Thoughts &53.90 &10.80 &26.34 &55.60 &\underline{41.63} &73.63 &43.65 &$4.99\times10^{17}$ \\
    MCTS &\underline{60.12} &10.80 &22.77 &\underline{56.80} &40.71 &\underline{74.74} & \underline{44.32} &$9.33\times10^{17}$ \\
    Guided Decoding &53.90 &10.80 &\underline{27.46} &53.20 &36.71 &73.55 &42.60 &$7.03\times10^{17}$ \\
    Predictive Decoding &58.00 &11.00 &22.10 &54.20 &39.78 &73.55 &43.11 &$4.73\times10^{17}$ \\
    \midrule
    \ours &\textbf{60.42} &\textbf{16.40} &\textbf{29.24} &\textbf{58.20} &\textbf{43.01} &\textbf{78.16} &\textbf{47.57} &$3.55\times10^{17}$\\
    \bottomrule
\end{tabular}
}
\caption{Main results. The optimal results are highlighted in bold, whereas suboptimal results are underlined. The \emph{Avg.} column indicates the averaged results across the six benchmarks. \emph{FLOPS} denotes the calculated computational cost, with lower values indicating lower costs.}
\label{exp_main}
\end{table*}

\paragraph{\ours significantly enhances the average performances of backbone LLMs.}
% \hangyan{may be better not to write "by 7-14\%"}}
Compared with the standard CoT strategy,
\ours can achieve the inference-time optimization without extra training. 
Specifically, notable average improvements of 14.62\% and 6.92\%  are observed in LLaMA3.1-Instruct and Mistral-v0.3-Instruct models respectively.

\paragraph{\ours strikes a superior balance between effectiveness and efficiency over strong baselines.}
In general, \ours outperforms the four strong baselines by a large margin, with consistent lower computational cost.
Compared with the recent promising MCTS-style method,
\ours showcases a notable average improvement of 3.25-5.70\%, achieved with one-third of the cost.
When contrasted with the recent SOTA baseline \emph{Predictive Decoding},
\ours shows remarkable superiority particularly in its adeptness at generalizing across various backbone LLMs.

\subsection{On the Inference-Time Scaling}

\begin{figure}[t]
\large
\centering
\includegraphics[scale=0.45]{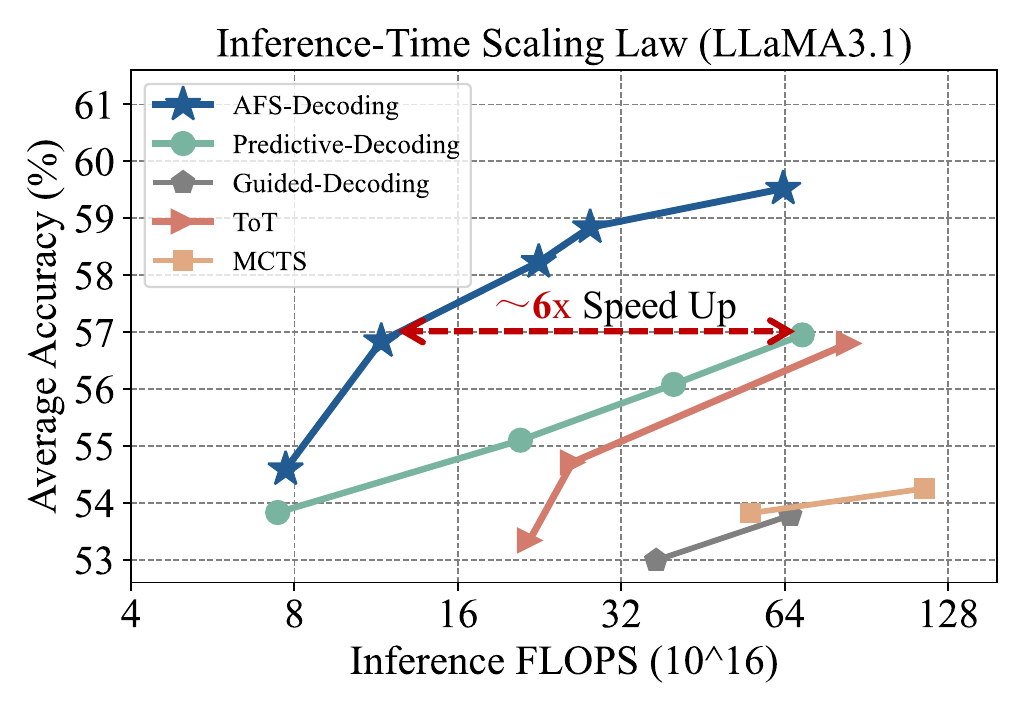}
\caption{Inference-time scaling law on LLaMA3.1-8B-Instruct. The horizontal axis denotes the inference-time computational cost, while the vertical axis represents the average performances on 6 benchmarks.}
\label{scaling}
% \vspace{-1em}
\end{figure}

Figure~\ref{scaling} presents the inference-time scaling law on LLaMA3.1-8B-Instruct.
From the scaling curves, \ours presents the consistent superiority on each computational budget, ranging from $8\times10^{16}$ to $64\times10^{16}$ FLOPS.
Furthermore, when considering similar performance levels (e.g., an average performance of $\sim$ 57\%), \ours demonstrates over 6$\times$ efficiency compared to even suboptimal methods.
Meanwhile, it is observed that \emph{Predictive Decoding} and \emph{ToT} also exhibit the stable improvement trend with the inference cost increasing.

%% file: sections/5.analysis.tex
\section{Analysis}

\begin{table*}[t]
\centering
\footnotesize
\resizebox{\linewidth}{!}{
\begin{tabular}{l|cccccc|cr}
    \toprule
    \multirow{1}{*}{\textbf{Models}}  &\textbf{GSM8K} &\textbf{Math-500} &\textbf{GPQA} &\textbf{ReClor} &\textbf{LogiQA} &\textbf{ARC-c} &\multirow{1}{*}{\textbf{Avg.}} &\multirow{1}{*}{\textbf{FLOPS}} \\
    \midrule
    \multicolumn{9}{c}{\cellcolor{gray!25} LLaMA3.1-8B-Instruct}  \\
    \midrule
    \ours &\textbf{86.58} &\textbf{38.20} &\textbf{34.60} &\textbf{64.00} &\textbf{48.39} &\textbf{85.41} &\textbf{59.53} &$6.43\times10^{17}$\\
    \quad w/o \emph{foresight sampling} &81.80 &35.00 &30.58 &60.60 &46.39 &84.90 &56.55 &$1.27\times10^{17}$\\
    \quad w/o \emph{cluster} &85.60 &37.40 &30.58 &61.00 &45.47 &85.32 &57.56 &$6.37\times10^{17}$\\ 
    \quad w/o \emph{dynamic pruning} &86.35 &38.20 &29.46 &61.00 &46.39 &85.67 &57.85 &$8.00\times10^{17}$\\
    \midrule
    \multicolumn{9}{c}{\cellcolor{gray!25} Mistral-v0.3-7B-Instruct}  \\
    \midrule
    \ours &\textbf{60.42} &\textbf{16.40} &\textbf{29.24} &\textbf{58.20} &\textbf{43.01} &\textbf{78.16} &\textbf{47.57} &$3.55\times10^{17}$ \\
    \quad w/o \emph{foresight sampling} &57.54 &11.40 &25.22 &42.40 &36.70 &75.60 &41.48 &$1.19\times10^{17}$\\
    \quad w/o \emph{cluster} &60.19 &15.00 &29.24 &56.60 &42.24 &76.45 &46.62 &$3.55\times10^{17}$\\
    \quad w/o \emph{dynamic pruning} &59.97 &15.20 &26.56 &53.20 &36.41 &75.77 &44.52    &$6.41\times10^{17}$\\
    \bottomrule
\end{tabular}
}
\caption{Ablation Studies on LLaMA3.1-8B-Instruct and Mistral-v0.3-7B-Instruct models. \emph{w/o foresight sampling} ablates the simulation of future steps. \emph{w/o cluster} ablates the calculation of \emph{Alignment} value. \emph{w/o dynamic pruning} ablates both of the pruning strategies.}
\label{exp_ablate}
\end{table*}

% \subsection{On the Effect of Foresight Sampling}
\subsection{Ablation Studies}

Some key components of \ours are ablated to verify their contributions to the overall performances in Table~\ref{exp_ablate}.
\emph{w/o foresight sampling} indicates that the look-ahead process is ablated, relying solely on step uncertainty for sampling.
\emph{w/o cluster} denotes that we simply sample on the foresight uncertainty distribution without considering the cluster distribution. 
\emph{w/o dynamic pruning} means the breadth and depth pruning strategies are ablated.
We have the following findings.

\paragraph{Foresight sampling mitigates auto-regressive generation limitations with extra inference cost.}
As the basis of our sampling strategy, simulating the future steps brings remarkable performance gains (2.98\%-6.09\%).
It proves the finding that the short-sightedness of the standard auto-regressive language generation can be reduced by increasing the inference-time computation.

\paragraph{Cluster distribution is beneficial to the overall performances.}
As one of the contributions,
we incorporate the cluster of foresight steps to mitigate the unreliability of the accumulated generation probability.
The results demonstrate that the cluster can calibrate the sampling distribution, leading to 0.95\%-1.97\% average gains.

\paragraph{Dynamic pruning largely reduces the computational costs.}
It is observed that the dynamic pruning strategy provides obvious efficiency improvement from the metric of \emph{FLOPS}.
Also, the dynamic pruning strategy surprisingly enhances model performance by eliminating distractions from negative rollouts during sampling.

\subsection{Generalization and Scalability}
Next, we analyze the generalization and scalability of \ours to (i) larger backbone LLM; and (ii) competition-level benchmarks.

\paragraph{\ours still works when scaling to 70B model size.}
Figure~\ref{70b} shows the results on LLaMA3.1-70B-Instruct across four benchmarks.
The model performance is further enhanced with the proposed algorithm.
It uncovers the superior generalization capability of \ours.
Limited by space, we leave the discussion of smaller backbone LLM (i.e., Qwen2.5-3B-Inst.) for Appendix~\ref{ref:generalization}.
The experiments on the 3B-sized model also reflect the obvious advantages brought by \ours.
Across the 6 reasoning benchmarks, \ours improves the backbone LLM by 3.80\% in average.
Combining all these generalization experiments,
it is concluded that \ours works well with a wide size range of LLMs, showcasing the superiority.

% \begin{figure}[t]
% \large
% \centering
% \includegraphics[scale=0.39]{Figures/70b.pdf}
% \caption{Generalization analysis on LLaMA3.1-70B-Instruct.}
% \label{70b}
% % \vspace{-1em}
% \end{figure}

\begin{table}[t]
\centering
\footnotesize
\resizebox{\linewidth}{!}{
\begin{tabular}{l|ccc}
    \toprule
    \multirow{1}{*}{\textbf{Tasks}}  &\textbf{AR (CoT)} &\textbf{\ours} &$\Delta$ \\
    \midrule
    GSM8K &92.27 &94.31 &{\color{darkgreen} +2.04} \\
    MATH-500 &41.40 &44.80 &{\color{darkgreen} +3.40} \\
    ReClor &67.60 &84.80 &{\color{darkgreen} +17.20} \\
    LogiQA &51.00 &56.37 &{\color{darkgreen} +5.37} \\
    \bottomrule
\end{tabular}
}
\caption{Generalization experiments on LLaMA3.1-70B-Instruct. The improvements over Auto-Regressive (CoT) are reported in the last columnn.}
\label{70b}
\end{table}
\vspace{0.2cm}

\paragraph{Our inference-time optimization can scale to improve performances on the competition-level task even with the strongest reasoning LLM.}

\begin{table}[t]
\centering
\footnotesize
\resizebox{\linewidth}{!}{
\begin{tabular}{l|cc}
    \toprule
    \multirow{1}{*}{\textbf{Methods}}  &\textbf{AIME2024} &\textbf{$\Delta$} \\
    \midrule
    LLaMA3.1-8B-Instruct &9.17 &- \\
    \quad + \emph{Predictive Decoding} &13.33 &{\color{darkgreen} +4.16} \\
    \quad + \ours  &16.67 &{\color{darkgreen} +7.50} \\
    \midrule
    R1-Distill-LLaMA-8B &37.81 &- \\
    \quad + \emph{Predictive Decoding} &20.00 &-17.81 \\
    \quad + \ours  &46.67 &{\color{darkgreen} +8.86} \\
    \bottomrule
\end{tabular}
}
\caption{Results on AIME 2024. We compare \ours with Predictive-Decoding based on two backbone LLMs: LLaMA3.1-8B-Instruct and R1-Distill-LLaMA-8B.}
\label{aime}
\end{table}

Table~\ref{aime} shows the results on AIME 2024 benchmark.
In addition to LLaMA3.1-8B-Instruct. and Mistral-v0.3-7B-Instruct.,
we also incorporate the DeepSeek-R1 model, utilizing the R1-Distill-LLaMA-8B variant due to resource constraints.
Even based on a well-trained deep thinking model, \ours can still help push the upper boundary on the competition-level task.
Such findings are exciting and insightful to implement inference-time optimization aimed at addressing challenging problems with LLM.

\subsection{Accuracy of Step Value Estimation}
\label{analysis_step}
\begin{figure}[t]
    \centering
    \includegraphics[scale=0.56]{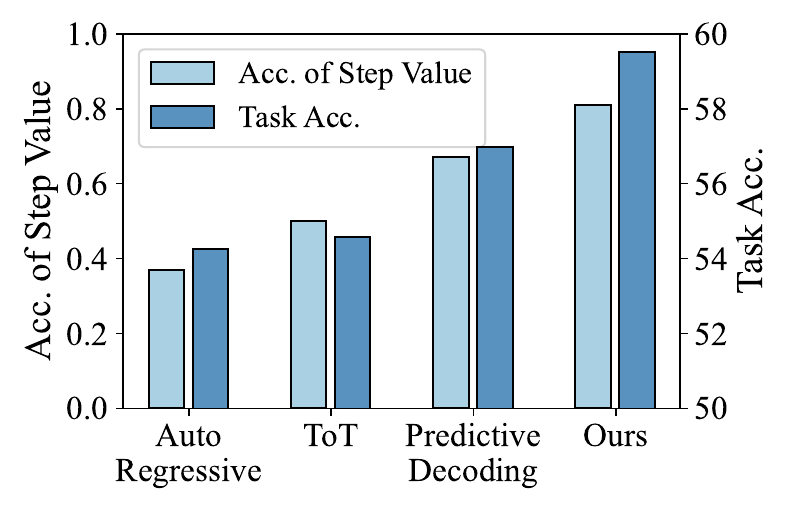}
    \vspace{-0.1cm}
    \caption{Analysis on the accuracy of step value estimation. The bar in light blue represents the accuracy of the step values, while the bar in dark blue denotes the averaged task performances.}
    \label{fig:step_value}
\end{figure}

The core of these decoding approaches is to estimate the precise step value through self-rewarding.
To measure how the estimated step value matches the actual rewards,
we employ the calculation of the \emph{Accuracy of Step Value} via distribution match.
Please refer to Appendix~\ref{ref:step_value} for details.
Based on the calculation, we visualize the results in Figure~\ref{fig:step_value}, revealing the following finding.

\begin{figure*}[t]
\large
\centering
\includegraphics[scale=0.35]{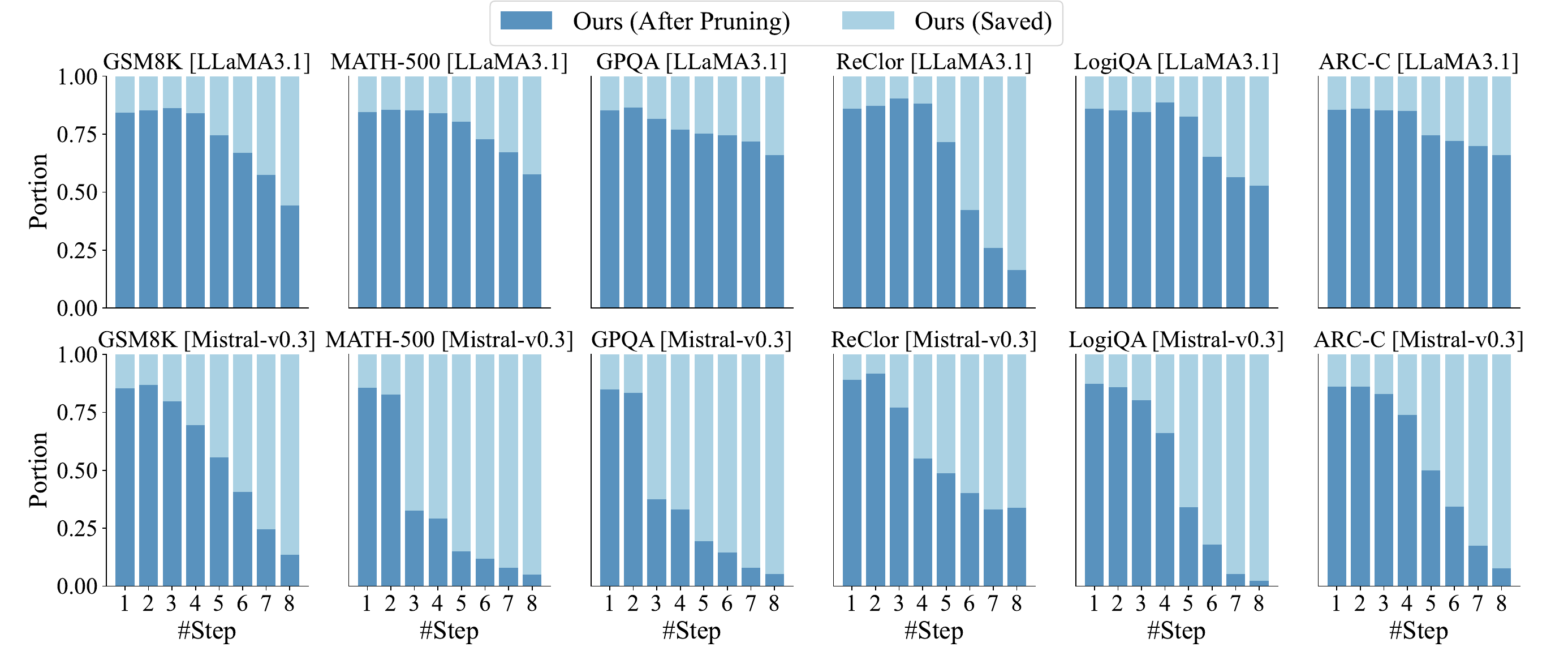}
\caption{Visualization of step-wise effects with alleviated overthinking. The first row displays the results for each independent benchmark using the LLaMA backbone, whereas the second row reflects the results with the Mistral backbone.}
\label{overthink}
\vspace{-1em}
\end{figure*}

\paragraph{The estimation of step value is positively correlated with the correctness of the final answer.}
Of the four inference-time decoding approaches illustrated in Figure~\ref{fig:step_value}, a more accurate estimation of the step value results in improved task performance.
Among them, \ours achieves the optimal estimation of step values as well as the final accuracy with obvious advantages.

\subsection{Analysis on Step-wise Overthinking}

Beyond simply reporting the \emph{FLOPS} metric,
a detailed analysis of the effects of pruning strategies is presented in Figure~\ref{overthink}.
It is observed that early steps are more critical, which involves relatively more computational costs.
At these early steps, it mainly relies on breadth pruning strategy to avoid redundant step exploration, reducing $\sim$ 20\% of the costs.
With the steps growing,
depth pruning takes over to alleviate overthinking through early stopping.
This finding inspires us to allocate more inference-time computational resources to the early steps, which are proved to be critical for the reasoning tasks.

%% file: sections/2.relatedwork.tex
\section{Related Works}

\paragraph{Inference-Time Optimization.}
To alleviate the post-training workload~\cite{zelikman2024quiet, liu2024acemath, qwq-32b-preview, guo2025deepseek}, inference-time optimization methods arouse wide concerns, showcasing a notable performance boost in reasoning scenarios~\cite{snell2024scaling, sun2023corex,zhao2024empowering}.
Mainstream methods can be categorized into searching-based~\citep{yao2024tree, hao2023reasoning, xie2024self,wu2024inference} and sampling-based~\cite{ma2024non, chen2023universal,zhang2024self}.
Although these works achieve the globally-optimal inference, they either induce large computation costs or yield inadequate step value estimation.
Other classical methods, such as Best-of-N, usually involve delegating the step selection to the external reward model~\cite{wang2024math, guan2025rstar}, and self-reflection strategies~\cite{cheng2024vision,xu2024interactive} usually involve extra training.
\ours stands out as an optimal and efficient decoding choice without reliance on external auxiliary.

\paragraph{Adaptive Inference-time Scaling.}
Though scaling of inference-time computations has proved to be effective~\cite{snell2024scaling},
the issue of over-thinking is widely observed and remains to be addressed~\cite{chen2024not}.
One line of works~\cite{team2025kimi, han2024token} stress on the control of the generation length,
while another line of methods~\cite{manvi2024adaptive, sun2024fast} leverage the idea of early-stopping.
In contrast, the adaptive scaling technique presented in our work is training-free and independent of external models.
Based on the self-evaluation of stepwise value, 
\ours introduces the comprehensive pruning strategy from the dimensions of width and depth.
It stands out as a light-weight solution to alleviate the inference-time over-thinking.

%% file: sections/6.conclusion.tex
\section{Conclusion}
This work focuses on inference-time optimization for LLMs, leveraging computational scaling to enhance performance. Building on stepwise reasoning and foresight sampling, we address two key research questions: (1) \emph{How can we achieve superior step value estimation?} and (2) \emph{Is deliberative planning necessary for every step?} 
We introduce a novel decoding strategy, \ours, that efficiently balances exploration and exploitation during inference. 
Extensive evaluations across seven diverse LLM benchmarks demonstrate \ours' state-of-the-art performance and efficiency. 
Furthermore, its ability to generalize to a wide range of LLMs (3B, 7B, 8B, and 70B) and scale across various computational budgets underscores the superiority of \ours in inference-time optimization.

% \section*{Limitations}

% (1) Due to limited computation resources, 
% we have not conducted experiments beyond reasoning tasks. 
% It is also interesting to evaluate the coding and agentic tasks (e.g., LiveCodeBench and AgentBoard), which we leave for future improvement.
% (2) In addition, our approach currently focuses solely on implementing pruning strategies to decrease computational costs during inference. Nevertheless, in complex and demanding scenarios, it may be more beneficial to dynamically increase the computational budgets to enhance deliberate reasoning.

%% file: sections/appendix.tex
\section{Implementation Details}

\subsection{Calculation of FLOPS}
\label{ref:flops}

Following~\cite{kaplan2020scaling}, we calculate the inference-time FLOPS to measure the computational efficiency.
The definition of the metric FLOPS is:
\begin{equation}
    \textrm{FLOPS} \approx 6nP
\end{equation}
where $n$ represents the total number of the output tokens, and $P$ is the number of parameters of the LLM.
In the tables above, we report the average results of FLOPS across the benchmarks.

\subsection{Inference Setup}
\label{ref:setup}

We provide the detailed implementation setup in Table~\ref{app:setup}.
Considering the huge cost ahead, the hyperparameters are merely searched within a very small range.
We leave it for future works to derive the optimal experimental configuration.

\begin{table}[h]
\centering
\footnotesize
\resizebox{\linewidth}{!}{
\begin{tabular}{l|ccccc}
    \toprule
    \textbf{Task} &\multicolumn{5}{c}{\textbf{Hyper-Parameter Setup}} \\
    \midrule
    \multicolumn{6}{c}{\cellcolor{gray!25} LLaMA3.1-8B-Instruct}  \\
    \midrule
    GSM8K &$M$=4 &$N$=4 &($T_{\mathrm{min}}$,$T_{\mathrm{max}}$)=(4,8) &$K$=3 &$\delta$=0.7 \\
    % GSM8K &\multicolumn{6}{c}{$M$=4, $N$=4, $T_{\mathrm{min}}$=4, $T_{\mathrm{max}}$=4, $K$=4, $\delta$=0.7} \\
    % GSM8K  &4 &4 &4 &8 &3 &0.7 \\
    MATH-500  &$M$=4 &$N$=4 &($T_{\mathrm{min}}$,$T_{\mathrm{max}}$)=(4,8) &$K$=3 &$\delta$=0.7 \\
    GPQA   &$M$=4 &$N$=4 &($T_{\mathrm{min}}$,$T_{\mathrm{max}}$)=(1,8) &$K$=3 &$\delta$=0.7\\
    ReClor  &$M$=4 &$N$=4 &($T_{\mathrm{min}}$,$T_{\mathrm{max}}$)=(4,8) &$K$=3 &$\delta$=0.7\\
    LogiQA &$M$=4 &$N$=4 &($T_{\mathrm{min}}$,$T_{\mathrm{max}}$)=(4,8) &$K$=3 &$\delta$=0.7\\
    ARC-C &$M$=4 &$N$=4 &($T_{\mathrm{min}}$,$T_{\mathrm{max}}$)=(4,8) &$K$=3 &$\delta$=0.7\\
    AIME2024 &$M$=3 &$N$=2 &($T_{\mathrm{min}}$,$T_{\mathrm{max}}$)=(32,64) &$K$=3 &$\delta$=0.7\\
    \midrule
    \multicolumn{6}{c}{\cellcolor{gray!25} Mistralv0.3-7B-Instruct}  \\
    \midrule
    GSM8K &$M$=4 &$N$=4 &($T_{\mathrm{min}}$,$T_{\mathrm{max}}$)=(2,8) &$K$=3 &$\delta$=0.7 \\
    MATH-500  &$M$=4 &$N$=4 &($T_{\mathrm{min}}$,$T_{\mathrm{max}}$)=(1,8) &$K$=3 &$\delta$=0.7\\
    GPQA  &$M$=4 &$N$=4 &($T_{\mathrm{min}}$,$T_{\mathrm{max}}$)=(1,8) &$K$=3 &$\delta$=0.7\\
    ReClor &$M$=4 &$N$=4 &($T_{\mathrm{min}}$,$T_{\mathrm{max}}$)=(2,8) &$K$=3 &$\delta$=0.7\\
    LogiQA &$M$=4 &$N$=4 &($T_{\mathrm{min}}$,$T_{\mathrm{max}}$)=(2,8) &$K$=3 &$\delta$=0.7\\
    ARC-C &$M$=4 &$N$=4 &($T_{\mathrm{min}}$,$T_{\mathrm{max}}$)=(2,8) &$K$=3 &$\delta$=0.7\\
    \midrule
    \multicolumn{6}{c}{\cellcolor{gray!25} Qwen2.5-3B-Instruct}  \\
    \midrule
    GSM8K  &$M$=4 &$N$=4 &($T_{\mathrm{min}}$,$T_{\mathrm{max}}$)=(4,8) &$K$=3 &$\delta$=0.7\\
    MATH-500  &$M$=4 &$N$=4 &($T_{\mathrm{min}}$,$T_{\mathrm{max}}$)=(4,8) &$K$=3 &$\delta$=0.7\\
    GPQA  &$M$=4 &$N$=4 &($T_{\mathrm{min}}$,$T_{\mathrm{max}}$)=(3,8) &$K$=3 &$\delta$=0.7\\
    ReClor &$M$=4 &$N$=4 &($T_{\mathrm{min}}$,$T_{\mathrm{max}}$)=(4,8) &$K$=3 &$\delta$=0.7\\
    LogiQA &$M$=4 &$N$=4 &($T_{\mathrm{min}}$,$T_{\mathrm{max}}$)=(4,8) &$K$=3 &$\delta$=0.7\\
    ARC-C &$M$=4 &$N$=4 &($T_{\mathrm{min}}$,$T_{\mathrm{max}}$)=(4,8) &$K$=3 &$\delta$=0.7\\
    \midrule
    \multicolumn{6}{c}{\cellcolor{gray!25} LLaMA3.1-70B-Instruct}  \\
    \midrule
    GSM8K  &$M$=4 &$N$=4 &($T_{\mathrm{min}}$,$T_{\mathrm{max}}$)=(7,8) &$K$=3 &$\delta$=0.7\\
    MATH-500  &$M$=4 &$N$=4 &($T_{\mathrm{min}}$,$T_{\mathrm{max}}$)=(3,8) &$K$=3 &$\delta$=0.7\\
    ReClor &$M$=4 &$N$=4 &($T_{\mathrm{min}}$,$T_{\mathrm{max}}$)=(2,8) &$K$=3 &$\delta$=0.7\\
    LogiQA &$M$=4 &$N$=4 &($T_{\mathrm{min}}$,$T_{\mathrm{max}}$)=(6,8) &$K$=3 &$\delta$=0.7\\
    \midrule
    \multicolumn{6}{c}{\cellcolor{gray!25} DeepSeek R1-Distill-LLaMA-8B}  \\
    \midrule
    AIME2024 &$M$=4 &$N$=4 &($T_{\mathrm{min}}$,$T_{\mathrm{max}}$)=(16,32) &$K$=3 &$\delta$=0.7\\
    \bottomrule
\end{tabular}}
\caption{Experimental setup of \ours. $M$ denotes the step beam size. $N$ is the number of rollouts for each step beam. $T_{\mathrm{min}}$ and $T_{\mathrm{max}}$ represent the least and the most foresight step number respectively. $K$ is the number of clusters while $\delta$ means the early-stopping threshold using clustering.}
\label{app:setup}
\end{table}

\FloatBarrier
\begin{table*}[t]
% \begin{table*}[H]
\centering
\footnotesize
\resizebox{\linewidth}{!}{
\begin{tabular}{l|cccccc|cr}
    \toprule
    \multirow{1}{*}{\textbf{Cluster Methods}}  &\textbf{GSM8K} &\textbf{Math-500} &\textbf{GPQA} &\textbf{ReClor} &\textbf{LogiQA} &\textbf{ARC-c} &\multirow{1}{*}{\textbf{Avg.}} &\multirow{1}{*}{\textbf{FLOPS}} \\
    \midrule
    \multicolumn{9}{c}{\cellcolor{gray!25} \ours (LLaMA3.1-8B-Instruct)}  \\
    \midrule
    TF-IDF &\textbf{86.58} &38.20 &\textbf{34.60} &\textbf{64.00} &\textbf{48.39} &\textbf{85.41} &\textbf{59.53} &$6.43\times10^{17}$\\
    SBERT (109M) &86.43 &\textbf{39.20} &33.26 &63.20 &47.48 &\textbf{85.41} &59.16 &$6.52\times10^{17}$\\
    SBERT (22.7M) &86.05 &36.80 &33.26 &62.40 &45.47 &\textbf{85.41} &58.23 &$6.61\times10^{17}$\\
    \bottomrule
\end{tabular}
}
\caption{Variants of cluster strategies.}
\label{app:cluster_strategy_1}
\end{table*}

\begin{table*}[htb]
\centering
\footnotesize
\resizebox{\linewidth}{!}{
\begin{tabular}{cc|cccccc|cr}
    \toprule
    \multirow{1}{*}{\textbf{$K$}} &\multirow{1}{*}{\textbf{$\sigma$}} &\textbf{GSM8K} &\textbf{Math-500} &\textbf{GPQA} &\textbf{ReClor} &\textbf{LogiQA} &\textbf{ARC-c} &\multirow{1}{*}{\textbf{Avg.}} &\multirow{1}{*}{\textbf{FLOPS}} \\
    \midrule
    \multicolumn{10}{c}{\cellcolor{gray!25} \ours (LLaMA3.1-8B-Instruct)}  \\
    \midrule
    3 & 0.7 &\textbf{86.58} &38.20 &\textbf{34.60} &64.00 &\textbf{48.39} &\textbf{85.41} &\textbf{59.53} &$6.43\times10^{17}$ \\
    2 & 0.8 & 85.52 & \textbf{39.40} &33.04 &\textbf{64.20} &46.85 &\textbf{85.41} &59.07 &$6.26\times10^{17}$\\
    4 & 0.5 & 83.93 &38.20 &32.37 &64.00 &43.78 &84.81 &57.85 &$6.15\times10^{17}$\\
    \bottomrule
\end{tabular}
}
\caption{Various setups of cluster.}
\label{app:cluster_strategy_2}
\end{table*}

\section{Algorithm of \ours}
\label{ref:algo}

\begin{figure}[h]
\large
\centering
\includegraphics[scale=0.6]{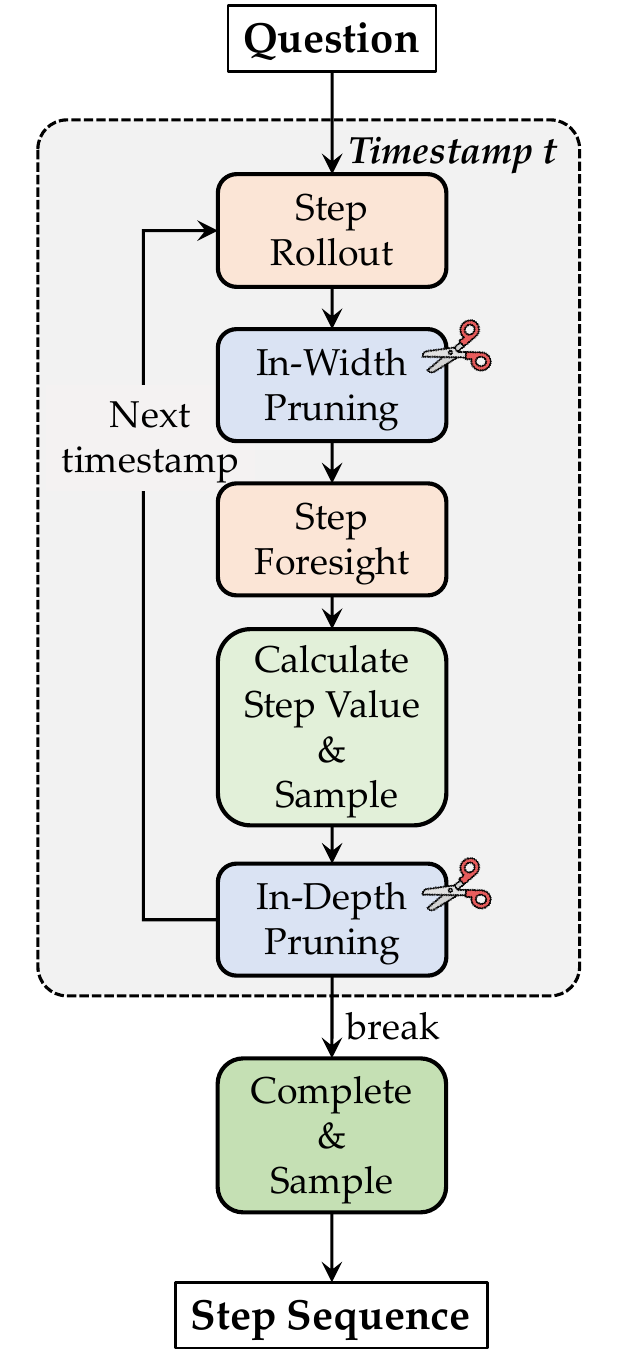}
\caption{Overall pipeline of \ours.}
\label{pipeline}
% \vspace{-1em}
\end{figure}

The pseudo code of \ours is presented in Algorithm~\ref{app:algo}.
To make a high-level overview of \ours, we also provide the pipeline in Figure~\ref{pipeline}.

\begin{algorithm*}
\caption{\ours}
\label{app:algo}
\begin{algorithmic}
\State {\bfseries Input:} Input query $x$, LLM $\pi_\theta$, step beam size $M$, number of rollouts on each beam $N$, minimum and maximum number of step foresight $T_{\textrm{min}}$ and $T_{\textrm{min}}$, number of clusters $K$, early-stopping threshold $\delta$.

\State {\bfseries Output:} Step sequence.
\State
% \State Set $\mathscr{S}_0 \gets$ Initialize $\texttt{env}$, $\texttt{finish}\gets$ False

% \For{$m = 1,...,M$}
%     \State $a_0^{(m)} \leftarrow empty \ string$
% \EndFor
% \State \hangyan{I added the above for loop, which makes the algorithm easier to understand}

% \State Initialize $a_0$ with empty string
% \hangyan{because $a_t$ will not appear in out algorithm, but $a_t^m$ will appear, I think it's better to do the following:}
% \For{$m = 1,...,M$}
%     \State$a_0^{(m)} \leftarrow $ empty\ string
% \EndFor
\For{$t = 1,2,\dots, T_{\textrm{max}}$}
    \State \Comment{Step Rollout (\emph{In Parallel})}
    \For{$m = 1,...,M$}
        \For{$n = 1,2,...,N$}
            \State Sample single step $a_t^{(m,n)}, s_t^{(m,n)} \sim p_\theta(\cdot|x,\mathbf{a}_{<t}^{(m)})$ 
            % \hangyan{should be $p_\theta(\cdot|x,\mathbf{a}_{<t}^{(m)})$}
        \EndFor
    \EndFor

    \State \Comment{In-Width Pruning (filter erroneous candidates)}
    % \State Derive mean and variance of step confidence: $\mu_t \leftarrow \frac{1}{M*N} \sum_{i} s_t^{(i)}, \, \, \sigma_t^2 \leftarrow \frac{1}{M*N} \sum_{i} (s_t^{(i)} - \mu_t)^2$

    % \State \hangyan{I think in the above statement, $i$ may confuse the readers with $(m,n)$, so you can change it to the following statement. (Although you have explained this in method part, it may still confuse some readers if they haven't fully understood our method)}
    \State Derive mean $\mu_t$ and variance $\sigma_t^2$ from these step confidence $s_t$
    
    \State Prune steps and keep the remaining ones for foresight: $\mathscr{S}_t \leftarrow \{a_t^{(m,n)} | \mu_t - \sigma_t \leq s_t^{(m,n)} \}$
    % \hangyan{better change $S$ to another letter, because it feels like the set of step confidence $s$ rather than $a$. (If you do so, you should also change the letter in method part)}
    
    \State \Comment{Step Foresight (\emph{In Parallel})}

    \For{each $a_t^{(m,n)}$ in $\mathscr{S}_t$} 
        \State Derive foresight steps and foresight scores: $\mathbf{a}_{>t}^{(m,n)}, F_t^{(m,n)}  \sim p_\theta(\cdot |x, \mathbf{a}_{\leq t}^{(m,n)})$,
    \EndFor

    \State \Comment{Step Value Esitimation (\emph{In Parallel})}
    \For{$i = 1, 2, \dots, |\mathscr{S}_t|$}
        \State $m,n \leftarrow$ the superscript of $i^{th}$ candidate in $\mathscr{S}_t$
        % \State Let $a_t^{(m,n)} \leftarrow \mathscr{S}_t[i]$
        % \State \hangyan{giving a value to $a_t^{(m,n)}$ is a bit weird here, consider change it to : $m,n \leftarrow$ the \ superscript\ of\ $i^{th}$ candidate in $S_t$}
        \State Derive \emph{Advantage} via $\Delta F$ of adjacent steps: $A_t^{(m,n)} \leftarrow F_{t}^{(m,n)} - F_{t-1}^{(m)}$
        
        \State Derive \emph{Alignment} via clustering: $C_t^{(m,n)} \leftarrow \texttt{Cluster}(\{\mathbf{a}_{>t}^{(m,n)}\})$
        
        \State Combine \emph{Advantage} and \emph{Alignment}: $R(x, \mathbf{a}_{\leq t}^{(m,n)}, \mathbf{a}_{>t}^{(m,n)}) \leftarrow \mathrm{Norm}(A_t^{(m,n)}) + \mathrm{Norm}(C_t^{(m,n)})$

        \State $w_i \leftarrow \mathrm{exp} \left[ R(x, \mathbf{a}_{\leq t}^{(m,n)}, \mathbf{a}_{>t}^{(m,n)}) / \tau \right]$
    \EndFor

    \State \Comment{Sample $M$ Steps}
    % $\hat{a}_t \sim p_\theta(a_t|x,\mathbf{a}_{<t}) \mathrm{exp} \left[ R(x, \mathbf{a}_{\leq t}, \mathbf{a}_{>t}) / \tau \right]$
    \For{$m = 1,2,...,M$}
        \State Sample without replace: $i \sim \textrm{Categorical}(\{\frac{w_i}{\sum_j w_j}\}_{i=1}^{|\mathscr{S}_t|})$
        \State Sampled step: $a_t^{(m)} \leftarrow \mathscr{S}_t[i]$ 
    \EndFor
    \State \Comment{In-Depth Pruning (early-stop)}
    \State \textbf{break} \textbf{if} $t \geq T_{\textrm{min}}$ and $\texttt{EarlyStop}(\delta) \text{ is True}$;

   %  \State Input \texttt{prompt} and $\mathscr{S}_0^{\prime}, a_0^{\prime},\dots,\mathscr{S}_t^{\prime}$ to language model \texttt{context};
   %  \For{$k = 1,2,\dots, K$} // \textcolor{gray}{ In parallel}
   %      \State Sample $a_t^k,\mathscr{S}_{t+1}^k, a_{t+1}^k,\dots, a_{t+T_0}^k, \mathscr{S}_{t+T_0+1}\sim P^\text{LLM}(\cdot \mid \texttt{context})$;
   %      \State $P_k \gets P^\text{LLM}(\mathbf{a}_{\geq t
   %      }^k,\mathscr{S}_{> t
   %      }^k  \mid \texttt{context})$;
   %      \State $w_k \gets \exp{\left(P_k/\tau\right)}$;
   %  \EndFor
   %  \State \Comment{Re-sample based on foresight.}
   %  \State Sample $j \sim \text { Categorical }\left(\frac{w_1}{\sum_{k=1}^K w_k}, \cdots, \frac{w_K}{\sum_{k=1}^K w_k}\right)$;
   %  \State Set $a_t^{\prime} \gets a_t^j$;
   %  \State \Comment{Takes the action $a_t^{\prime}$}
   %  \State Update $\mathscr{S}_{t+1}, \texttt{finish}\gets$ Execute $a_t^{\prime}$ in environment $\texttt{env}$;
   % \State \textbf{break} \textbf{if} $\texttt{finish} \text{ is True}$;
\EndFor
\State Complete all candidates at the last foresight step and sample only one based on the $R$ function.
\State \textbf{Return} Step sequence.
\end{algorithmic}
\end{algorithm*}

\section{Generalization to Smaller LLMs}
\label{ref:generalization}

Besides the generalization to 70B-sized backbone, we also supplement the evaluations on 3B-sized model.
Table~\ref{app:qwen3b} presents the performances on Qwen2.5-3B-Instruct model.

\begin{table}[h]
\centering
\footnotesize
\resizebox{\linewidth}{!}{
\begin{tabular}{l|ccc}
    \toprule
    \textbf{Tasks} &\textbf{AR(CoT)} &\textbf{\ours} &$\Delta$  \\
    \midrule
    \multicolumn{4}{c}{\cellcolor{gray!25} Qwen2.5-3B-Instruct}  \\
    \midrule
    GSM8K &78.62 &85.60 &{\color{darkgreen}+6.98}\\
    MATH-500 &41.00 &45.20 &{\color{darkgreen}+4.20}\\
    GPQA &28.57 &28.79 &{\color{darkgreen}+0.22} \\
    ReClor &53.60 & 59.40 &{\color{darkgreen}+5.80}\\
    LogiQA &42.70 &46.08 &{\color{darkgreen}+3.38}\\
    ARC-C &77.47 &79.69 &{\color{darkgreen}+2.22}\\
    \midrule
    \textbf{Avg.} &53.66 &57.46 &{\color{darkgreen}+3.80} \\
    \bottomrule
\end{tabular}
}
\caption{Generalization to smaller backbone Qwen2.5-3B-Instruct.}
\label{app:qwen3b}
\end{table}

Compared with the auto-regressive chain-of-thought baseline,
\ours provides obvious performance gains across 6 reasoning benchmarks, improving the average performance by 3.80\%.

\section{Accuracy of Step Value Estimation}
\label{ref:step_value}

To measure whether the estimated step value aligns with the actual rewards, we conduct the analysis in Sec.~\ref{analysis_step}.
At each timestamp $t$, we can derive the value estimation of the candidate steps via the decoding strategy.
These step values can approximate a distribution $P_1$.
Meanwhile, we can derive the explicit outcome of each candidate step using the foresight paths.
Comparing the outcome with ground-truth, the outcome accuracy for these candidate steps can also form a distribution $P_2$, where $|P_1| = |P_2|$.
We derive the distribution matching as the accuracy of step value estimation:
\begin{equation}
    \textrm{Accuracy} = 1-\frac{\sum_{i=1}^{|P_1|} {(P_1(i) - P_2(i))}^2}{|P_1|}
\end{equation}
where  $P_{1}(i) \in P_1$,  $P_{2}(i) \in P_2$. 
In the implementation of $P_1$, we use the model estimated step values for sampling-based methods (\ours and Predictive Decoding). 
For auto-regressive and ToT methods,
we allocate binary rewards for the selected steps (rewarded as 1) and other candidates (rewarded as  0).
The final accuracy score is calculated by averaging the results on each timestamp.

\section{In-depth Analysis of Cluster Strategies}

\subsection{Variants of Cluster}
\label{ref:cluster_variants}

In the main experiments, we implement the cluster strategy with TF-IDF, which is from the syntax perspective.
It can also be replaced with sentence-BERT (SBERT)~\cite{reimers-2019-sentence-bert} to obtain the sentence embedding for clustering.

Table~\ref{app:cluster_strategy_1} presents the comparisons between different cluster strategies.
SBERT (109M) employs the pretrained sentence embedding model of $\texttt{multi-qa-mpnet-base-dot-v1}$,
while SBERT (22.7M) utilizes the model of $\texttt{all-MiniLM-L6-v2}$.

From the results, clustering with the external embedding model can also lead to similar competitive performances, slightly lower than the TF-IDF strategy.
Also, it is observed that increasing the size of the sentence embedding models can bring improvements in the average performances.

\subsection{Hyperparameters of Cluster}
\label{ref:hyper}

Table~\ref{app:cluster_strategy_2} offers the analysis on different hyper-parameters.
We keep the other configuration fixed for fair comparison, where $M$=4 and $N$=4.
Under this setting, the maximum number of foresight paths for clustering is 16.
Based on the results, the cluster size $K$=2 or 3 would be good choices.
With $K$ increasing, it may bring much uncertainty.

% \section{Case Study}
% Figure~\ref{case_study} demonstrates an example of solution trajectory on GSM8K. In this case, $M$ is set to 2, $N$ is set to 2, $T_{min}$ is set to 2 and $T_{max}$ is set to 4. 

% During each timestamp, \ours firstly rollout $M*N=4$ candidate steps. Using in-width pruning, we then prune some steps with relatively low step confidence (denoted in gray). After that, we conduct foresight on remaining candidates (denoted in orange) to generate step value and sample $M=2$ candidate steps based on it.
% After $T_{max}$ = 4 timestamps, we complete the remaining $M=2$ candidates and select only one final answer among them.

% It's worthy to note that we observe no in-depth pruning in this case, due to the complexity of the question. Besides, in each timestamp, we evaluate the correctness of foresight answers generated from each  candidate. The average accuracy in each timestamp improves gradually (0, 0, 0.33, 0.67, respectively), which reveals that \ours can continuously lead to a better solution space.

% \begin{figure*}[t]
% \large
% \centering
% \includegraphics[scale=1]{Figures/case_study.pdf}
% \caption{Case study}
% \label{case_study}
% % \vspace{-1em}
% \end{figure*}

% \section{Case Study}